# RAPID: A Reachable Anytime Planner for Imprecisely-sensed Domains

**Emma Brunskill**
Computer Science Department
University of California, Berkeley
Berkeley, CA

**Stuart Russell**
Computer Science Department
University of California, Berkeley
Berkeley, CA

## Abstract

Despite the intractability of generic optimal partially observable Markov decision process planning, there exist important problems that have highly structured models. Previous researchers have used this insight to construct more efficient algorithms for factored domains, and for domains with topological structure in the flat state dynamics model. In our work, motivated by findings from the education community relevant to automated tutoring, we consider problems that exhibit a form of topological structure in the factored dynamics model. Our Reachable Anytime Planner for Imprecisely-sensed Domains (RAPID) leverages this structure to efficiently compute a good initial envelope of reachable states under the optimal MDP policy in time linear in the number of state variables. RAPID performs partially-observable planning over the limited envelope of states, and slowly expands the state space considered as time allows. RAPID performs well on a large tutoring-inspired problem simulation with 122 state variables, corresponding to a flat state space of over $10^{30}$ states.

## 1 INTRODUCTION

One of the key questions in artificial intelligence research is how to make good decisions in large, stochastic, partially observable environments. Though generic optimal planning for finite-horizon partially observable Markov decision processes (POMDPs) is known to be PSPACE-complete (Papadimitriou & Tsitsiklis, 1987), fortunately, some important POMDP domains have highly structured models. This insight has been used by previous researchers to design more efficient POMDP algorithms that leverage different types of structure. Focussing on domains that exhibit factored structure has led to POMDP planners that solve some of the largest POMDP problems in the literature, including a hand washing assistance program (Boger et al., 2005) and a RoboCup rescue task (Paquet et al., 2005). Other recent work (Dai & Goldsmith, 2007; Dibangoye et al., 2009) has focused on domains where the flat state dynamics model limits the possible backtracking to earlier states, and showed that planning can be performed more efficiently when this topological structure is present.

In this paper we focus on problems exhibiting both factored structure and a form of topological structure, and demonstrate that we can leverage these properties to scale to very large domains. Such properties are common in a number of important applications ranging from tutoring to dialogue systems. For example, some prior education studies coarsely approximate a student's knowledge as a factored set of binary variables, one for each skill, and infers a precondition graph structure among skills (known as a "learning hierarchy") from student data: see for example Gagneé's and Briggs (1974) and Close and Murtagh (1986). Despite this structure, automated tutor action selection remains challenging as the factored state space may consist of hundreds of skills. In addition, the student state is not directly observable, but can be probed through the use of drill exercises and other student responses. Modelling a fairly small curriculum of 100 skills using an atomic-state POMDP framework could require planning over a state space of size $2^{100} \approx 10^{30}$ which is far outside the range of generic, flat POMDP solvers.

Specifically we consider constructing policies for POMDPs that exhibit the following three properties: they are

1. *factored*,
2. have *positive-only* effects, and
3. have *unique preconditions* for each variable.

For compactness, in the rest of the paper we will refer to Positive-Only effects, Factored, with Unique Preconditions (POFUP) POMDPs as POFUPP processes. Factored representations are those in which the world state is represented by a vector of variables. Positive-only effects, commonly leveraged in classical planning, imply that once a binary variable becomes true, it will not later become false. Before we describe the third property, recall that in a factored representation, a given state variable $s^k$'s value on a subsequent time step depends on the action chosen, and the values of a set of the other state variables (which could include $s^k$ on the previous time slice): in a dynamic Bayes net (DBN), these would be called the parents of $s^k$. The unique preconditions assumption implies that there is a single set of values of $s^k$'s precondition variables that allow $s^k$ to become true. In all the education learning hierarchies we examined, there was always a unique set of preconditions for each variable. It is important to note that while there is a unique set of preconditions for each state *variable*, there are still numerous (potentially exponential in the number of variables) paths to reach each *state*. We assume that the planning objective is to reach a goal state.

Our Reachable Anytime Planner for Imprecisely-sensed Domains (RAPID) leverages these three structural properties to construct an initial policy with a computational cost that scales polynomially with the number of domain variables, instead of exponentially. RAPID first computes a solution to the fully observable MDP starting at an initial state sampled from the initial POMDP belief state. This process is very fast, taking only time linear in the number of state variables. RAPID then performs partially-observable planning over the limited envelope of states reached under this MDP policy, and then slowly expands the state space considered as time allows. At most the state space envelope will expand to become the reachable state space given the initial potential starting states, which is typically much smaller than the exponential potential state space.

We present promising experimental results on two large tutoring-inspired simulations. The second problem consists of 122 variables, or a potential flat state space of over $10^{30}$. RAPID manages to achieve good performance quickly in both problems, though several comparison planners, including a factored approach, fail to find a good policy.

## 2 RELATED WORK

There has been significant recent progress on planning in partially observable, stochastic domains. Two of the fastest generic POMDP planners are HSVI by Smith and Simmons (2005) and SARSOP by Kurniawati, Hsu and Lee (2008). Neither approach takes advantage of factored structure.

A number of prior fully-observable MDP approaches do leverage factored structure (such as Boutlier, Dearden and Goldsmidt (2000)). Symbolic Perseus (Boger et al., 2005) and Symbolic HSVI (Sim et al., 2008) are two offline POMDP algorithms for factored state spaces which scale to large problems. In practice both perform fairly similarly to each other. Online, forward search POMDP planners can also leverage factored structure, and Paquet, Tobin and Chaib-draa (2005) used forward search to handle an extremely large, factored RoboCup rescue problem. However, their approach and other forward search techniques typically scale as $O((|A||Z|)^H)$ where $H$ is the search horizon,and $|A|$ and $|Z|$ are, respectively, the action and observation branching factors. Such approaches will typically struggle in long horizon problems with a large number of observations or actions unless value heuristics can be used to shape the search. Unlike our algorithm, these factored approaches do not leverage any further structure in the domain dynamics.

Several recent approaches do seek to leverage topological structure in the dynamics model similar to the structure implied by our second and third assumptions. Dai and Goldsmith (2007) leveraged the presence of layered positive-effect state structure (certain clusters of states cannot be returned to) in their Topological Value Iteration (TVI) MDP algorithm. Dibangoye et al. (2009) assumed a similar structure, and used this to create a heuristic Topological Order Planner (TOP) for POMDPs. These and related approaches consider structure in the ground state space: in contrast, our approach considers structure in the factored space. Focussing on structure in the factored space helps our approach to scale to large domains as we can often avoid even enumerating the flat state space.

Finally, our approach is inspired by work in the fully observable planning community. To scale to very large, fully observable MDPs, Dean et al. (1995) proposed an anytime approach which initially restricts MDP planning to a smaller envelope of reachable states. Gardiol and Kaelbling (2004) extended this approach to be applicable in relational MDPs using action-based equivalence. To our knowledge our RAPID algorithm is the first approach that performs envelope-based planning in partially observable environments.

## 3 PROBLEM DESCRIPTION

We are interested in decision making in POFUP partially observable, stochastic environments that may be specified by the tuple $\langle S, L, A, Z, b_0, E, p((s^i)'|s^i, a), \ldots p(z|(s^i)', a), r(s,a), s_G, s_T \rangle$ where

- $S$ is a set of states. The domain consists of $L$ binary-valued variables $s^1, s^2, \ldots, s^L$, and each state is an assignment of values (true or false) to all the domain variables: $s = \langle s^1, s^2, \ldots, s^L \rangle$.
- $A$ is a set of actions. Each action $a_{ij}$ is associated with a particular state variable $s^i$ and has the potential to make only that variable true.[1] There will generally be multiple actions associated with the same state variable $s^i$. For example, there could be a drill exercise

[1]Actions or operators which have a single effect have been previously described as unary operators (Brafman & Domshlak, 2003).

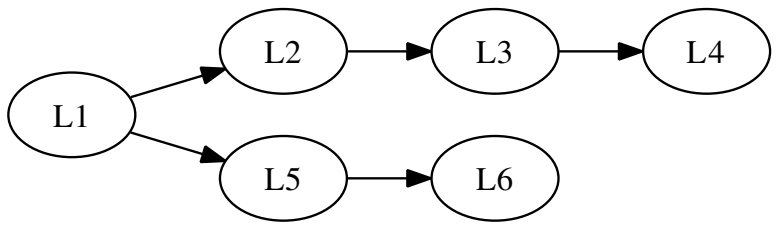

(a) Precondition graph

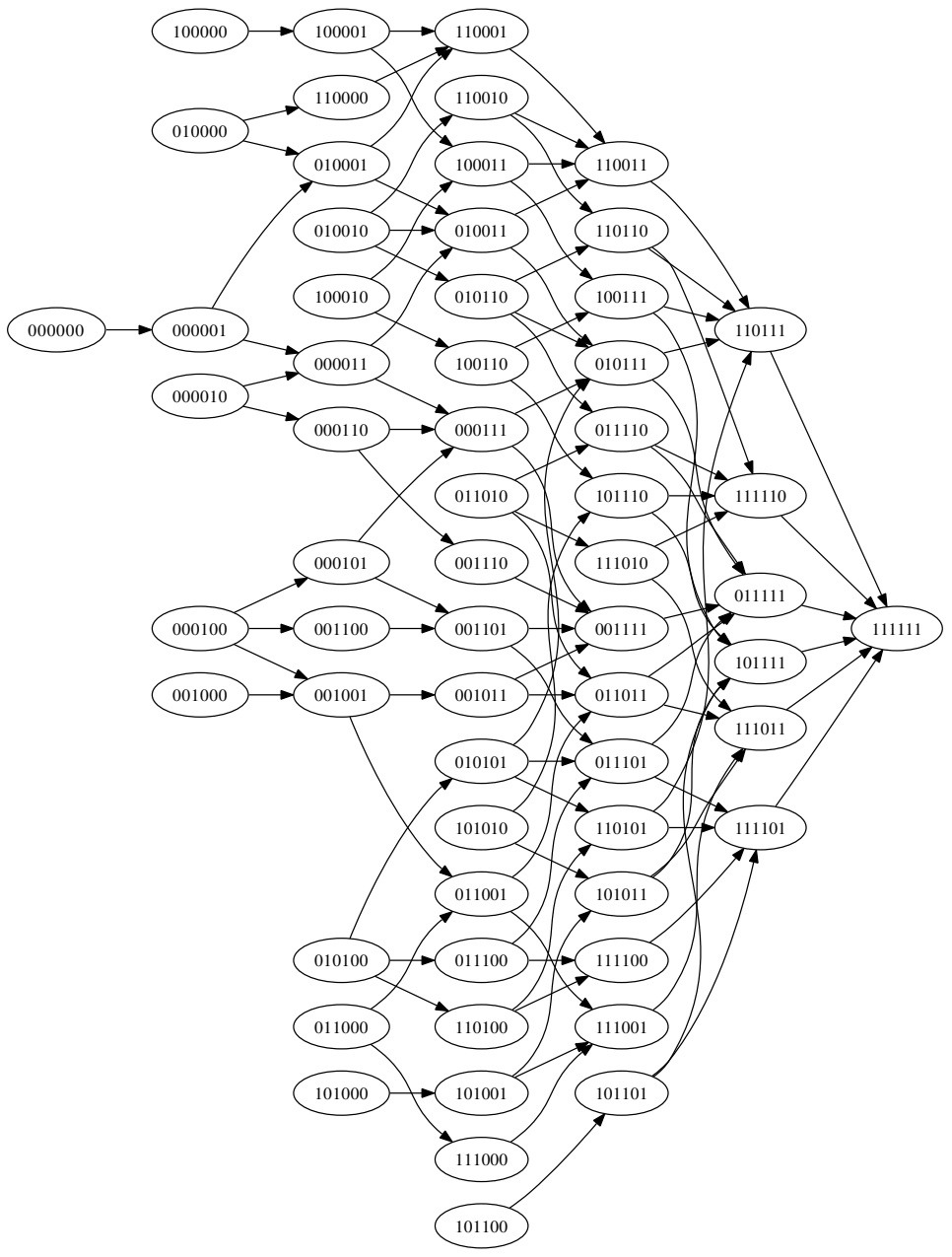

(b) State space

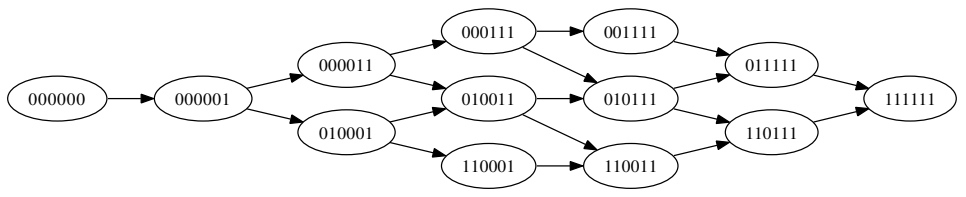

(c) States reachable from $s_0 = \{0, 0, 0, 0, 0, 0\}$

Figure 1: The relationship between the precondition variable graph of 6 binary state variables, the possible state space and transitions, and the reachable state space starting at a particular initial state.

action and a lesson action to help a student understand two-digit addition.

- $Z$ is a set of observations.
- $b_0$ is the initial belief state which is a sparse representation of the possible initial states and associated probabilities. The sum of the probabilities over all possible initial states is constrained to equal 1.
- $E$ is a precondition graph which specifies for each state variable $s^i$ the set of state variables $s^{ip1}, s^{ip2}, \ldots s^{ipM}$ (equivalent to parents of $s^i$ in a DBN) that must be true before state variable $s^i$ can become true. We assume there is a unique conjunction of precondition variables for each variable (for example, $s^1 \wedge s^2$ can be a precondition, but not $s^1 \vee s^2$). As a concrete example, the precondition graph for a student to master the multiplication skill would include the addition skill as a prerequisite.
- $p((s^i)' = false | s^i = false, a_{ij})$ specifies the probability of a state variable $s^i$ remaining false even when all $s^i$'s preconditions are satisfied and a relevant action $a_{i*}$ is taken. If a state variable $s^i$'s preconditions are not satisfied, and action $a_{i*}$ is applied, $s^i$ always remains false. Continuing the prior example, let $a_{ij}$ be a multiplication exercise, and $s^i$ be the multiplication skill. Then $p((s^i)' = false | s^i = false, a_{ij})$ is the probability that after trying a multiplication exercise, a student still may not yet understand multiplication, even if she has all the necessary preconditions skills (addition, etc.) as specified in $E$.
- $p(z|(s^i)', a_{ij})$ specifies the probability of receiving a particular observation given that a particular action $a_{ij}$ is taken, and the resulting value of the action's associated state variable $(s^i)'$. Note that since an action is only associated with a single variable, only a single $p(z|(s^i)', a_{ij})$ will be applicable at each time step.
- $r(s, a_{ij})$ is the reward for taking action $a_{ij}$ in state $s$. The reward is negative, and depends only on the action (aka independent of the state) for all states except the goal state $s_G$ and terminal state $s_T$.
- $s_G$ is the goal state. $r(s_G, a)$ is positive or zero.
- $s_T$ is the terminal state. $s_G$ deterministically transitions to $s_T$. $s_T$ is a sink state where the reward is 0 and the observation probabilities are identical to the observation probabilities of $s_G$.

Figure 1 shows the relation between the number of variables as expressed in a precondition graph, and the potential state space and state transition graph.

As the states are partially observable, we maintain a distribution over states, known as the belief state, which is a sufficient statistic of the history of actions taken and observations received. The planning objective is to maximize the expected sum of rewards given the initial belief state $b_0$. Due to the reward formulation, this is similar to a partially observable, stochastic shortest path problem.

# 4 ALGORITHM

Prior flat and factored POMDP approaches typically fail to scale to domains with a large number of variables. This often continues to hold true even when, for particular initial belief states, the reachable state space is significantly smaller than the full state space.

Instead we draw inspiration from envelope-based planning algorithms for large fully observable MDPs and extend these ideas to our POFUPP domains. Dean et al. (1995) presented the idea of computing a policy for fully observable, flat MDPs by planning only over a smaller envelope of states. As time allowed, the state envelope was expanded to include more of the reachable state space.

**Algorithm 1** RAPID: REACHABLE ANYTIME PLANNING FOR IMPRECISELY-SENSED DOMAINS

1: Sample an initial state from the initial belief
2: Construct an initial envelope using a deterministic MDP relaxation that can be solved efficiently.
3: **while** remaining time **do**
4: Define & solve a POMDP over the envelope
5: Expand the envelope
6: **end while**

To our knowledge RAPID is the first algorithm to take a similar approach in the context of partially-observable planning. There are several key technical challenges that need to be overcome to apply envelope-based planning in partially observable domains that can be characterized as POFUPP problems. First, we require an algorithm for efficiently computing a good initial envelope over the large, factored, partially observable state space. Second, we need a method for converting this envelope into a fully defined POMDP and solving the resulting model. We present solutions for both these challenges, and RAPID's empirical efficiency allows us to scale to very large problem sizes. The RAPID algorithm is summarized in Algorithm 1.

### 4.1 INITIAL ENVELOPE CONSTRUCTION

Given a POFUPP process $M$, we first need to construct an initial envelope of states. Ideally the envelope would include states that have a reasonable probability of being visited given a good policy for the partially observable domain. The states visited along the optimal MDP solution starting with one of the possible initial states would seem intuitively to be reasonable, as the MDP solution forms an upper bound on the POMDP performance. However, standard MDP value iteration will be intractable since it scales as a function of the state space, which in our process is an exponential function of the number of variables. Even alternate factored solvers will typically be too slow.

Instead we propose an approach which leverages the particular properties of our structured process by first relaxing the process to its deterministic, fully observable equivalent, and use this to very quickly compute a good trajectory between a start state $s_0$ and the goal $s_G$.

We first sample a state $s_0$ from the initial belief state $b_0$. Given $s_0$, and the variable precondition graph $E$, RAPID identifies the state variables whose value is false in $s_0$ and true in the goal state $s_G$. RAPID then computes a topological order of these state variables given the precondition graph $E$. A topological order of these variables is any linear ordering such that each state variable comes before all other state variables to which it has outbound arrows in the precondition graph. For example, in Figure 1a, state variable L1 must appear before all other variables, and L2 must appear before L3. As the precondition graph $E$ is a directed acyclic graph (DAG)[2], the topological order can be computed in time linear in the number of state variables and precondition conditions (Cormen et al., 1999).

The computed topological state variable ordering (such as $\langle s^2, s^{68}, \ldots s^{16} \rangle$) is converted into a state trajectory between the start $s_0$ and goal state $s_G$ by simply adding in order each state variable to the original $s_0$. Therefore the cost of generating an initial envelope is simply a linear function of the number of state variables. In Section 4.5 we will show that this state trajectory consists of the state variables visited by following an optimal MDP policy for $M$ starting at the sampled state $s_0$.

### 4.2 ENVELOPE POMDP POLICY GENERATION

RAPID proceeds by defining a POMDP $P'$ over the current state envelope. We supplement the envelope state space defined by the state trajectory sequence by two additional states: a terminal out state $s_{tout}$, and a terminal goal state $s_{tg}$. The definition of an out state follows prior work in the fully observable envelope literature (Dean et al., 1995). The dynamics of the states within the envelope are the same as in the original process $M$, except if a state transition lead to a state outside the envelope, then that transition, and associated probability, are set to go to the $s_{out}$ state. The out state itself transitions with probability one to the terminal out sink state $s_{tout}$ which has self-loop dynamics. The separation of $s_{tout}$ and $s_{out}$ is done in order to specify separate reward functions.

To discourage leaving the envelope, the reward for the out state is set to a large negative value. $s_{tout}$ has reward zero. Separating $s_{out}$ from $s_{tout}$ allows there to be a single shot cost for exiting the envelope.[3]

The observation model for all states within the envelope is the same as in the POFUPP $M$. In contrast to envelope planners for fully observable MDPs where all states, including the out state, is fully observed, in POMDP domains the out states are most naturally modeled as partially observable, since they represent the remaining partially observable states that are not in the envelope. This raises the interesting side problem of how to represent the observation probabilities for the out states, which represent the potential observation probabilities of all states outside the envelope. In general there will be an exponential (in the number of variables) states outside of the envelope, and so for now we take the simple approach of approximating the observation probability of $s_{out}$ by averaging the observation models of a sampled set of states lying outside the envelope. The observation model of $s_{tout}$ is identical to $s_{out}$.

[2] Since actions have positive-only effects, there are also no cycles in the corresponding state dynamics.

[3] An alternate strategy would be to define rewards over state, action, next state tuples.

If there is any initial probability over states outside of the envelope, then a new belief state is defined over only the envelope state space, with all remaining probability mass in the out state $s_{out}$.

POMDP $P'$ can be solved using any generic POMDP planner with optimality bounds and in our experiments we used the publicly-available HSVI (Smith & Simmons, 2005). POMDP planning proceeds until the error bound over the initial belief state drops below a chosen $\epsilon$-threshold, or a specified time limit is reached.

Note that the computed policy for POMDP $P'$ can be used to act in the original POFUPP $M$.

### 4.3 ENVELOPE EXTENSION

If additional planning time is available after the initial policy is computed, then the state envelope can be expanded. There are numerous potential strategies for envelope expansion and in this initial work we used a simple, but empirically effective approach. We consider three possible methods, in order, for identifying a new state to add to the envelope; in other words, we try the first method and see if it identifies a new state to be added, if it does, we stop, else we run the second method, etc.

The first method samples any potential initial state $s_{0i}$ which has non-zero probability in the initial belief state $b_0$, but is not yet part of the envelope of states. If all potential initial states are in the envelope, the second method tries to find a new non-envelope state by expanding the envelope fringe. This expansion is performed by starting at a possible initial state and simulating a trajectory using an $\epsilon_r$-greedy policy[4] until either a non-envelope state is reached, or a goal state is reached. This process is repeated until a non-envelope state is reached or a set number of iterations pass. If no non-envelope states are found, in the third method, we iterates through each state and tries all applicable actions (given the preconditions the state represents) to see if a new non-envelope state is reachable. This ensures that, given enough time, the envelope will grow to reach the full reachable state space, given the possible initial states defined by the initial belief state.

Once a non-envelope state is identified, it must be added to the envelope. In many cases these newly-added states will be multiple state transitions from the existing envelope of states. For example, consider a mathematics tutor domain where to start a student either knows algebra, or algebra and calculus. If the initial envelope is constructed starting from the state where the student knows calculus, and then the state representing the student only knows algebra is added, there are many missing steps between algebra and calculus that need to be added in order to compute a reasonable policy for the newly added initial state. To address this, when a new potential state is added, RAPID re-performs the initial envelope construction of creating a complete state trajectory to the goal, starting from the newly added state. This process is very fast, and the main limitation of this approach is that it can add $O(L)$ states to the envelope per state, which slows down the POMDP planning process. However, the benefit of increasing the probability that the new states will immediately improve the computed policy, was thought to outweigh this slight shortcoming.

[4]The POMDP policy is followed $(1-\epsilon_r)$ fraction of time time, and a random action is taken $\epsilon_r$ fraction of the time.

### 4.4 PERFORMANCE AND COMPUTATIONAL COMPLEXITY

First, for completeness, we note that RAPID is guaranteed to converge to an $\epsilon$-optimal policy, as long as an $\epsilon$-optimal POMDP planner is used, since RAPID is guaranteed to eventually expand the envelope to include all states reachable from the initial belief state.

Computing a state trajectory from an initial to goal state, and associated value computations, takes time linear in the number of variables. The initial envelope will have at most $O(L)$ states, which means that the initial POMDP planning will be performed over a state space which is a linear function of the number of variables. The maximum number of states in the envelope is the reachable state space, which is typically much smaller than the potential $2^L$ state space. The complexity of solving a POMDP depends on the particular technique. HSVI performs a depth-first roll out, and updates an explicit representation of an upper and lower bounds on the POMDP value function along the roll out. Each lower bound backup and belief update is a quadratic function of the number of states, so both operations will be impacted positively by a smaller input state space.

### 4.5 UPPER BOUNDS FOR POFUPP PROBLEMS

We will shortly prove that the trajectory of states between a start and the goal state, as computed during envelope initialization and expansion, consists of states visited by following an optimal policy for the fully observable MDP of the POFUPP process. We leverage this property to efficiently compute the fully-observable optimal MDP value of the states within the envelope, which can then be used to calculate an upper bound on the initial belief state $b_0$. Such bounds can be useful for at least two reasons. First, many POMDP solvers (including HSVI and SARSOP) use upper bounds during planning. Typically these bounds are computed by solving the MDP, which is known to be an upper bound to the POMDP values. However, solving the flat MDP typically requires multiple backup operations, each of which requires time polynomial in the number of states. Second, upper bounds provide useful benchmarks for evaluating RAPID's performance. However, solving the MDP upper bound over the complete factored space of hundreds or more variables is computationally infeasible. In contrast, our approach scales as $O(LN_{b0})$ where $N_{b0}$ is the number of initial states with non-zero probabilities.

We now illustrate how we compute the value of the the states along a trajectory between a start and goal state, as returned during envelope initialization and expansion. We first modify the original rewards. Let $\tilde{r}(s_{-i}, a_{ij})$ be the new reward for taking action $a_{ij}$ in a state $s_{-i}$ where state variable $s^i$ is false but all its preconditions are true. We define the value of this new reward as:

$$\tilde{r}(s_{-i}, a_{ij}) = \frac{r(s_{-i}, a_{ij})}{1 - p((s_{-i})' = false | s_{-i} = false, a_{ij})}. \quad (1)$$

Intuitively, $\tilde{r}(s_{-i}, a_{ij})$ represents the expected reward/cost of making state variable $s^i$ true using action $a_{ij}$, given the stochasticity of action $a_{ij}$. To compute the state trajectory values, we start with the goal state, and traverse the trajectory backwards, at each step selecting the action $a_{ij}$ with the minimum expected cost $\tilde{r}$ required to make the subsequent variable $s^i$ in the consecutive state true. The values are computed simultaneously, by summing up the rewards during the traversal:

$$a^*(s_{-i}) = \underset{j}{\operatorname{argmax}}\, \tilde{r}(s_{-i}, a_{i,j}) \quad (2)$$

$$V(s_{-i}) = \tilde{r}(s_{-i}, a^*(s_{-i})) + V(s_{+i}) \quad (3)$$

where state $s_{+i}$ is the same as state $s_{-i}$ except now state variable $s^i$ is also true. This value computation requires time linear in the number of variables. This process can be done at the same time as when the state trajectory is constructed from the topological order.

**Theorem 1.** *Given a POFUPP $M$, let $M_f$ be the fully-observable MDP version of $M$, $s_0$ be a state sampled from $b_0$, $\{s_0, s_{traj1}, \ldots, s_G\}$ be the state trajectory computed by the initial envelope method, $\pi_{M_f}$ be the associated policy, and $V(s_0), \ldots, V(s_G)$ be the calculated state trajectory values. These values and policy represent an optimal policy and the optimal values of these states in the MDP $M_f$.*

*Proof.* (Sketch) The initial topological order constructed is an optimal plan to the goal from the start state $s_0$ for the deterministic, uniform action-cost, fully observable process $M_{duf}$ version of the POFUPP $M$. This is true due to the particular POFUPP structure assumed. Briefly, the positive-only effects and the presence of unique preconditions to make a single variable true, imply that all permutations (that respect the precondition structure) of the same set of state variables will result in the same final state. As in $M_{duf}$ all rewards are constant except at the goal, all paths of the same length between the same start state and the goal state will have the same cost. Therefore we can arbitrarily select any ordering that respects the preconditions, and its value is guaranteed to be optimal (and equal to all other topological orderings between the same start state and goal).

To determine the optimal value (and policy) of each state along the corresponding state trajectory in the deterministic (but with the original action costs/rewards) MDP $M_{df}$ version of $M$ requires considering the state-action values of each state. From Bellman the state-action value can be expressed as the immediate reward of taking an action in a state, plus the future expected reward. As we currently assume each action is deterministic, the state action value of a state $s_{-k}$ which is a state where state variable $s^k$ is false but all its preconditions are true, can be expressed as

$$Q(s_{-k}, a_{kj}) = \tilde{r}(s_{-k}, a_{kj}) + V(s_{+k}) \quad (4)$$

where $s_{+k}$ is the state identical to $s_{-k}$ except state variable $s^k$ is also true. In the deterministic MDP $M$, $Q(s, a_{kj})$ represents the expected cost of making the state variables in $s_G$ true which are false in the current state $s$. However, since in a POFUPP process each variable requires a unique set of precondition variables to be true, the order in which these state variables are acquired is irrelevant: any order that satisfies the precondition structure $E$ is equivalent. The only difference in rewards/costs comes from which action $a_{kj}$ out of a set of actions $a_{k*}$ is chosen to achieve a state variable $s^k$; note here that all $a_{k*}$ have the same preconditions, but they may have different costs, different self-transition probabilities, and different observation probabilities. Therefore, the state trajectory obtained from the topological order is equal to any other trajectory of states between $s_0$ and the goal $G$. Given this, action selection for each state along the trajectory can be restricted without loss of optimality to only those actions which pertain to the next variable to be acquired along the topological order (as specified by Equation 2). This means that the next state $s_{+k}$ in Equation 4 will be identical for all considered actions $a_{k*}$, and to find the optimal action it suffices to only consider the immediate expected reward $\tilde{r}$. Therefore the policy and values in Equation 2 and 3, respectively, represent an optimal policy and the optimal value for the deterministic MDP.

Finally, the MDP $M_f$ falls into the class of stochastic shortest path problems. Therefore the computed value function and policy for the deterministic MDP $M_{df}$ which has no-self loops (as just specified) using the modified rewards defined in Equation 1 has an identical policy and value function to the original MDP $M_f$ (pg.25, Bertsekas and Tsitsiklis, (1996)). Therefore, the values and policy computed using Equation 2 and 3 for all states along the trajectory are guaranteed to be optimal for MDP $M_f$. □

Theorem 1 shows that we can efficiently compute the optimal MDP values for states inside the constructed envelope.

Once the envelope includes all possible initial states, we will have a value on each initial state computed using Equation 3. We can then compute an upper bound for initial belief state value ($\bar{V}(b_0)$) by taking the weighted sum of the

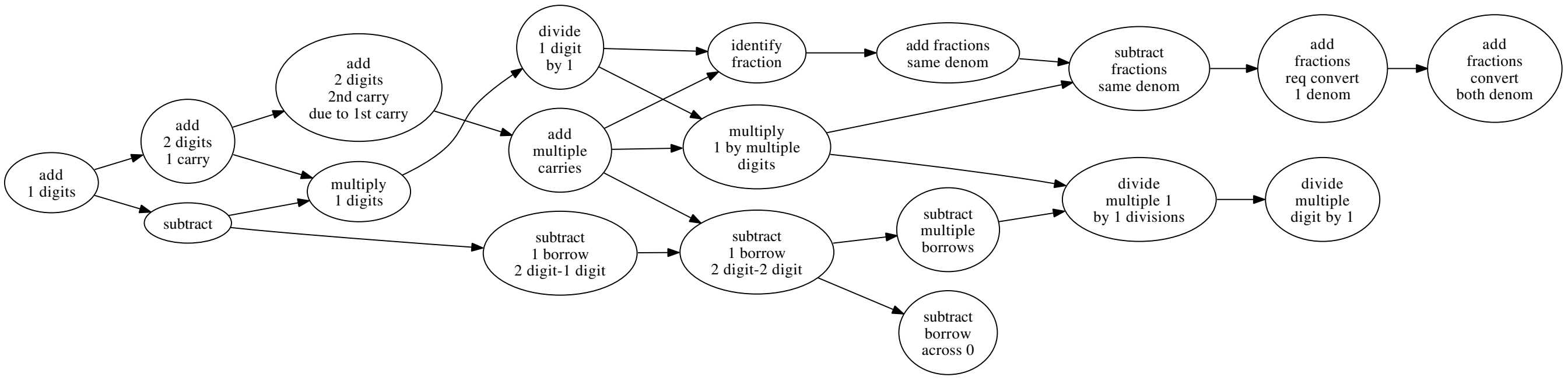

Figure 2: SmallMath Precondition Graph.

state values:

$$\bar{V}(b_0) = \sum_{s_{si}} b_0(s_{si})V(s_{si}). \tag{5}$$

Note that this provides an upper bound to the original PO-FUPP process: in contrast, any bounds computed by the POMDP solvers over the envelope only apply to the restricted envelope POMDP $P'$. We later calculate $\bar{V}(b_0)$ for our two experimental domains. This bound can be computed in time linear in the product of the number of state variables and initial possible states.

# 5 EXPERIMENTS

Due to our interest in tutoring applications, we performed simulation experiments in two tutoring-inspired domains.

## 5.1 DOMAINS

In both cases, the variable precondition graph construction was informed by literature from the education communities: the transition probabilities, observation probabilities and reward values were chosen by hand.

The first domain, SmallMath, consisted of 19 elementary math skills, yielding a potential state space of $2^{19} \sim 500,000$ states. The precondition graph for the skills is displayed in Figure 2. There are two possible observations, and 38 actions, 2 for each skill. The first action for a skill, a "teaching" action, has a high probability of causing the skill to transition to being true ($p = 0.8$) if it is not already and the preconditions for that skill are fulfilled; however, it does not provide any feedback about whether the student has successfully acquired the skill. In our experiments we set the probability of each observation is $0.5$ for actions 1,3,5,. . .,37. The second action for each skill (actions 2,4,. . .,38) loosely corresponds to a practice exercise, and only causes skill acquisition with probability $0.5$. However, practice exercises provide more useful feedback about whether that skill was acquired: the observation is true with probability 0.9 if the hidden skill is true, and true with probability 0.2 if the skill is false. The reward for reaching the state where all skills are true was set to 10000, and there was a reward of -1 for all other states and actions. The initial belief state had three non-zero initial start states.

In the past there have been a number of papers on "learning hierarchies" in the education literature. Learning hierarchies consist of ordered hierarchies, or graphs, of skills, which are very similar to our variable precondition graphs. Numerous classroom studies have been done to construct these learning hierarchies from student data, though the analysis historically treats the data as fully observable rather than modeling student knowledge as a hidden state.

Given this work, for the second domain, BigMath, we constructed a larger tutoring-inspired problem consisting of addition, subtraction, multiplication, and addition and subtraction of fractions skills. The fraction precondition graph was derived from Miller and Phillips (1974) and Uprichard and Phillips (1977). We combined the fraction precondition structure with the subtraction hierarchy from Gagné (1974), and the addition, subtraction, multiplication and division hierarchies from Close and Murtagh (1986). The full precondition graph is displayed in Figure 3 and consisted of 122 skills.[5] The flat state space is $2^{122}$ which is over $10^{30}$ states. Similar to the first domain, we created an action space with two potential actions for each skill, one lesson-like action, and one drill-like action. The observation and transition probabilities, given the precondition variables are satisfied, were defined the same way as in the SmallMath domain. The reward for reaching the state where all skills are true was set at 100000, and there was a reward of -1 for all other states and actions. The original belief state had four non-zero probability initial start states, consisting of plausible variable subgroups.

Note that the horizon of both problems is quite long. Even in the deterministic versions of both problems, if the world state starts with no variables true, the number of steps to reach the goal is 19 in SmallMath and 122 in BigMath.

[5] A file displaying this precondition structure is available at http://www.cs.berkeley.edu/∼emma/bigmathpreconditions.pdf

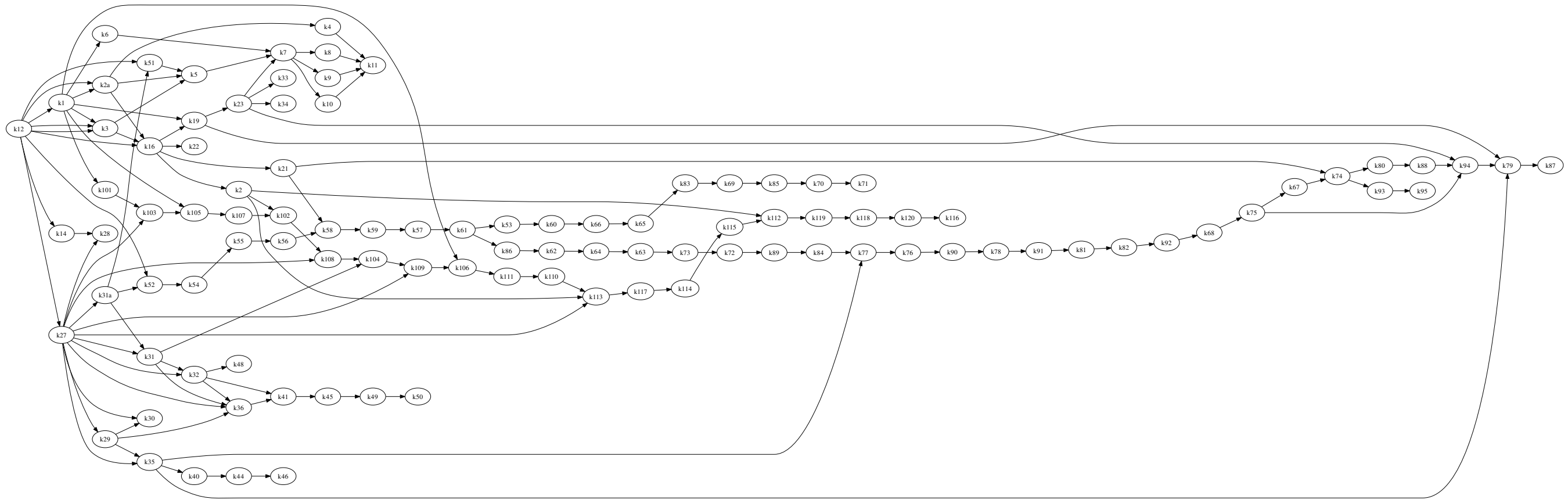

Figure 3: Precondition Graph for BigMath. This structure was derived by combining the Miller and Phillips (1974) and Uprichard and Phillips (1977) learning hierarchies for fractions, with Gagneé and Briggs' (1974) subtraction hierarchy and Close and Murtagh's (1986) addition, subtraction, multiplication and division hierarchy.

As both problems are stochastic, the expected number of steps can be significantly longer, depending on the initial belief state distribution. Therefore, both domains exhibit what are typically known in the POMDP community as the curse of history, due to the long horizon, and the curse of dimensionality, due to the problem size.

### 5.2 SOLUTION PARAMETERS

As stated earlier, we used HSVI to solve the envelope POMDPs. The maximum horizon for SmallMath was set at a conservative 450 steps, and for BigMath at 1000 steps. Identical horizon limits were used when evaluating the empirical reward of the computed policy. The reward for the out state was set to be -1000 for SmallMath and -100 for BigMath. As there will typically be some probability that the state will transition into an out-of-envelope state, and both problems can require a long horizon of acting to reach the goal, the out state reward was loosely chosen to discourage transitioning to the out state without so severely penalizing the transition that the computed policy conservatively avoids adding any more skills. We did not optimize performance by varying this parameter, and other values might lead to further performance benefits.

HSVI terminates when a terminal time limit is reached or a minimum distance ($\epsilon$) between the upper and lower bounds on value of the initial belief state is achieved. In SmallMath we set the maximum time limit to 1200 seconds and $\epsilon = 200$. In BigMath we set the maximum time limit to 8000 seconds and $\epsilon = 1000$.

### 5.3 EVALUATION METRICS

After each envelope expansion, we evaluated the envelope policy reward empirically over multiple episodes of the relevant problem's max horizon length. For SmallMath we evaluated the empirical reward for 20 episodes after each expansion, and for BigMath we evaluated the empirical reward for 5 episodes after each expansion: BigMath is significantly more computationally intensive to evaluate due to the larger state space, and longer problem horizon. We present results averaged over 5 runs with different initial seeds for SmallMath and 8 runs for BigMath.

### 5.4 BASELINES

Even the smaller of the two problems, SmallMath, still requires over 500,000 states to enumerate the exhaustive set of state variable combinations, which limited the potential alternate algorithms to compare against.

SARSOP (Kurniawati et al., 2008) is a non-factored state-of-the-art generic POMDP solver which accepts factored input files.

Symbolic Perseus (Boger et al., 2005) is a factored-state-space POMDP solver. Symbolic Perseus was used to compute a good approximate solution to a factored hand-washing assistance problem with 13 variables, and over $50 * 10^6$ states.

In some cases the reachable state space may be quite small, and so we also explored first enumerating the reachable space, and then using HSVI to compute a POMDP policy over the reachable states.

We also implemented a simple, very fast, heuristic Fixed Threshold, No-Forgetting (FTNF) policy similar to policies used in prior intelligent tutoring systems (Corbett & Anderson, 1995; Koedinger et al., 1997). At each step, FTNF identifies the variable with the highest probability of being true below an input threshold probability, whose preconditions have exceeded this threshold probability. FTNF executes the action most likely to make that variable true, and updates the belief probability over that variable. Once a variable exceeds the input probability threshold, its value is assumed to be true for the rest of the episode.

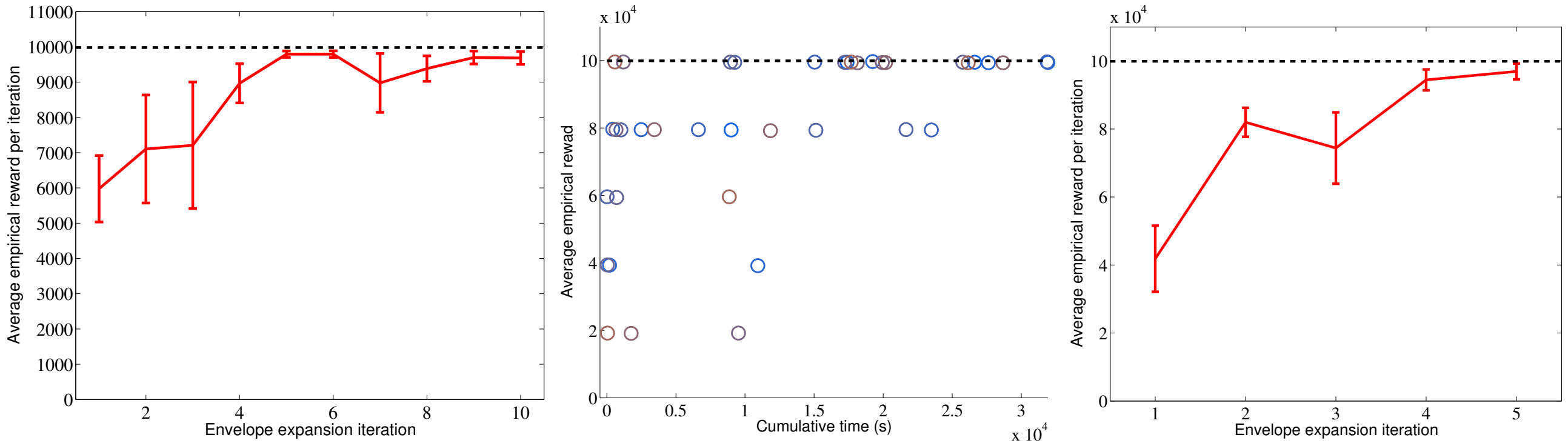

(a) SmallMath Reward vs No. Expansions (b) BigMath Reward vs. Cumulative Time (c) BigMath Reward vs. No. Expansions

Figure 4: Results from both simulations, showing RAPID's average performance after each round of envelope expansion. In (a) and (c) results are averaged over multiple algorithm runs. In (b) each RAPID run is shown in a different color, with circles representing the mean reward of the run's policy at different times. The dotted line is an upper bound on the initial belief state value.

Finally, we also computed an upper bound on the value of the initial belief state using Equation 5.

## 5.5 RESULTS

### 5.5.1 SmallMath

We display the performance of RAPID on SmallMath in Figure 4a. RAPID could generally quickly find a good solution, and its consistency in doing so increased as the computation time increased, as should be expected.

We represented SmallMath in the SARSOP POMDPX format but found that SARSOP problem initialization consistently tried to exceed our limit of available memory (2 gigabytes). We believe this is because the current implementation still uses a non-factored dynamics representation, and a full sized-representation of SmallMath would require $2^{19} \times 2^{19} \times 38$ entries.

Symbolic Perseus requires specifying the number of sampled belief states to use for planning. When we specified a small number of beliefs (N=20), the algorithm computed a solution in 5150s, but the resulting policy could never find a trajectory to the goal. Using a larger number of beliefs (N=120), Symbolic Perseus was still generating belief points and had yet started computing a policy after 8 hours; as this well exceeded the time necessary to achieve good performance in the SmallMath domain using RAPID, we did not run Symbolic Perseus further.

Given the initial belief selected, the reachable state space of SmallMath is significantly smaller than the potential state space size, at only 109 states. It is computationally tractable to simply enumerate this reachable state space and run HSVI over the resulting states. This approach yielded the best performance, with an average reward of 9962 on 200 trials (each consisting of at most 200 steps). This corresponded to an average of 39 steps to reach the goal state. The heuristic FTNF policy performed worse than the RAPID policy over a number of thresholds, and was significantly worse (t-test, p¡0.001) than the POMDP solution over the reachable state space at even the best threshold (0.925) examined (FTNF average reward=9947, mean number of steps to goal=54). These results highlight the advantage of a POMDP planning approach, which may both infer the value of earlier variables based on later variable values, and revisit an earlier variable if later evidence suggests its value is not yet true.

### 5.5.2 BigMath

RAPID again was able to fairly quickly achieve good performance in this domain. Figure 4b & c display the average performance after each envelope expansion for different runs versus cumulative running time, and after each envelope expansion, respectively.

Due to our experience with SARSOP and Symbolic Perseus on SmallMath, we did not explore their use on BigMath, which is a substantially larger problem.

In BigMath, given the chosen initial belief, even the reachable space is over millions of states and the potential state space exceeds $10^{30}$ states. It was therefore not feasible to perform standard planning over the reachable space.

We compared FTNF to the performance of RAPID after 4 envelope expansions. Though FTNF is very fast, it generally performed much worse than RAPID over a wide range of thresholds (from 0.8 to 0.9999). FTNF with the best found threshold (0.9999) performed slightly better than RAPID over an 80 episode simulation, but the difference was not significant (t-test, p=0.18). Our experience suggests that it may be hard to identify a good threshold for FTNF in advance, and choosing a too-high value can lead to overly conservative policies.

## 6 CONCLUSION & FUTURE WORK

There exist a number of important stochastic, partially observable problems that exhibit a large amount of structure that can be used to perform efficient planning. In this paper we focused on problems exhibiting a form of topological structure in the factored state space: domains which possess such structure include student tutoring, dialogue and potentially assembly tasks. Our RAPID algorithm leverages this structure to compute an initial state envelope based on the optimal MDP policy in time linear in the number of variables. RAPID then performs standard POMDP planning over this restricted envelope, before expanding the envelope and re-solving in an anytime fashion. Our experimental results demonstrate RAPID can quickly produce a good policy for an extremely large factored problem where the problem structure is constructed using prior precondition graphs from the education community.

There is ample scope for future work. We intend to explore additional envelope expansion techniques, such as trying to bias the new trajectories to the goal to lie within existing parts of the envelope. In addition, we currently re-solve the POMDP without considering the previously computed solution. We believe it should be possible to achieve further computational gains by re-using the value function ($\alpha$-vectors) computed using the prior envelope, by setting the value of the additional states to a lower bound on their potential value.[6] In this paper we have assumed the POMDP model parameters are provided, but to integrate this in a real ITS will necessitate learning the model parameters. We plan to learn model parameters across multiple students' performances, motivated by the success of prior ITSs (see Koedinger et al. 1997) that use population-level model parameters.

### Acknowledgements

The authors wish to thank Sarah Finney, Jason Wolfe, Luke Zettlemoyer and the anonymous reviewers for their helpful comments. E.Brunskill was supported by a NSF Mathematical Sciences Postdoctoral Fellowship.

[6] Trivial lower bounds can be computed using the minimum reward and maximum horizon.